\documentclass{article}
\usepackage{ijcai18}

\usepackage{times}
\usepackage{xcolor}
\usepackage{soul}
\usepackage[utf8]{inputenc}
\usepackage[small]{caption}
\usepackage{graphicx}
\usepackage[colorlinks]{hyperref}
\usepackage{tabularx}

\usepackage{balance}

\newcommand{\etal}{\textit{et al.\ }}
\newcommand{\ie}{\textit{i.e.}}
\newcommand{\eg}{\textit{e.g.}}

\title{Towards Automatic Identification of Elephants in the Wild}

\author{
Matthias K{\"o}rschens, 
Bj{\"o}rn Barz, 
Joachim Denzler 
\\ 
Computer Vision Group, \\ Friedrich Schiller University Jena \\ 
\{matthias.koerschens,bjoern.barz,joachim.denzler\}@uni-jena.de
}
\begin{document}

\maketitle

\begin{abstract}
	
	Identifying animals from a large group of possible individuals is very important for biodiversity monitoring and especially for collecting data on a small number of particularly interesting individuals, as these have to be identified first before this can be done. Identifying them can be a very time-consuming task. This is especially true, if the animals look very similar and have only a small number of distinctive features, like elephants do. In most cases the animals stay at one place only for a short period of time during which the animal needs to be identified for knowing whether it is important to collect new data on it.
	For this reason, a system supporting the researchers in identifying elephants to speed up this process would be of great benefit.
	
	In this paper, we present such a system for identifying elephants in the face of a large number of individuals with only few training images per individual. For that purpose, we combine object part localization, off-the-shelf CNN features, and support vector machine classification to provide field researches with proposals of possible individuals given new images of an elephant.
	
	The performance of our system is demonstrated on a dataset comprising a total of 2078 images of 276 individual elephants, where we achieve 56\% top-1 test accuracy and 80\% top-10 accuracy.
	
	To deal with occlusion, varying viewpoints, and different poses present in the dataset, we furthermore enable the analysts to provide the system with multiple images of the same elephant to be identified and aggregate confidence values generated by the classifier.
	With that, our system achieves a top-1 accuracy of 74\% and a top-10 accuracy of 88\% on the held-out test dataset.
	
\end{abstract}

\begin{figure}
	\centering
	\includegraphics[width=0.68\linewidth]{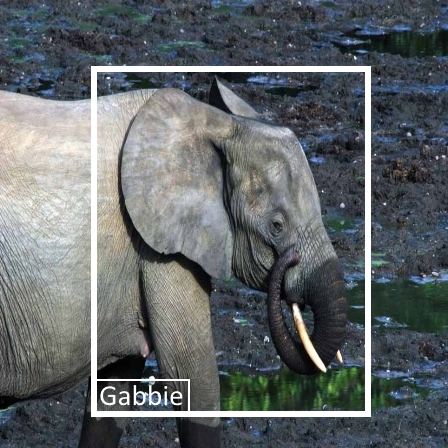}
	\caption{An example classification containing the automatically detected head bounding box and the predicted name of the elephant.}
	\label{fig:teaser}
\end{figure}

\section{Introduction}

In biological research projects, there often is a focus on certain individuals from a larger group. These individuals might be particularly interesting, be it because they are the matriarch of a huge family, or simply because they have a certain special behavior, not commonly found among others of its kind, and thus it is important to collect new data from them. As many of the individuals might look very similar, it can be very hard to identify them on the spot.

Sometimes researchers are staying with such animals for a long time, get familiar with them and thus also become able to identify them without any notes taken before. But as this is most often not the case, and most researchers often only stay there for shorter periods of time, they are not able to get that familiar with the animals and thus have to go through all the archived features concerning each individual for identification. This is not only a very time-consuming but also exhausting task.
Additionally, the need for manual identification can lead to stagnation of research, because the animals of interest may move to another place before they can be identified, so that valuable data that could have been collected about them is lost.

This can slow down the research progress significantly. Thus, it would be very advantageous to have a supportive system, which can help researchers to identify the individuals quickly. In this work we propose to achieve this by combining object localization for finding elephant heads, off-the-shelf CNN features as descriptors, and support vector machines (SVMs) for multi-class classification. An example identification of an elephant from our dataset is shown in \autoref{fig:teaser}.

The remainder of this paper is organized as follows: We briefly review related work in \autoref{struc:related_work} and describe our proposed system for identification of individual animals in \autoref{struc:methods}, after introducing the elephant dataset used for our work in \autoref{struc:dataset}. Experimental results are presented in \autoref{struc:results} and, after a short problem analysis in \autoref{struc:problems}, we will conclude this work in \autoref{struc:summary}.

\section{Related Work}
\label{struc:related_work}

In the context of human beings, face identification is a very actively studied field, where breakthroughs have recently been achieved using deep learning with systems trained end-to-end, \eg, FaceNet \cite{schroff2015facenet}, VGG-Face \cite{parkhi2015deep}, or DeepFace \cite{taigman2014deepface}.
However, such approaches usually require large amounts of annotated training images per class, which are often not available in wildlife monitoring scenarios.

Loos \etal hence used traditional face identification methods such as Eigenfaces and SVMs for identifying 25 individual chimpanzees \cite{loos2011identification} and later extended this by automatic face detection \cite{loos2013automated}.

Brust \etal have recently taken this approach to the deep learning age for gorilla identification using pre-trained convolutional neural networks (CNNs) for face detection and feature extraction \cite{Brust17_TAV}.
Their approach is, in principle, very similar to ours. However, we do not only demonstrate that it is also suitable for identifying other species such as elephants, but also show that the performance can be improved further by using earlier layers than the last layer of a CNN for feature extraction and additional pooling. Moreover, we found that simple data augmentation such as flipping can be useful for training the SVM classifier and show how to aggregate predictions obtained for multiple images of the same unknown individual to deal with occlusion and variations in pose and perspective.

In contrast to the formerly mentioned works, our dataset also poses new challenges: a large number of classes, a very small and imbalanced number of images per class, and a long time period during which images have been taken (17 years).

\section{Elephant Dataset}
\label{struc:dataset}

We use a dataset provided by biologists from Cornell University, who run research on elephants in the Kongo in a project called \textit{The Elephant Listening Project}\footnote{\url{http://www.elephantlisteningproject.org/}}, especially in the region of the Dzanga-Sangha special reserve \cite{turkalo2017slow}. This research has been going on since 1999, mostly focusing on one clearing many different elephants come to every year. 

Over the years, about 4000 different elephants were sighted there and documented by the researchers in the field.
They assigned names to many of the elephants and captured photos and videos of them manually. Distinctive features, which can be used to identify the elephants, were documented. Since those features can be very subtle or even change over time, identifying the individual elephants can be very hard.
A reliable and fast identification system would hence be of great benefit for their research.
%As a reliable and fast identification system would be important to have for their research, they provided the dataset used for our experiments.

\begin{figure}
	\centering
	\includegraphics[width=0.97\linewidth]{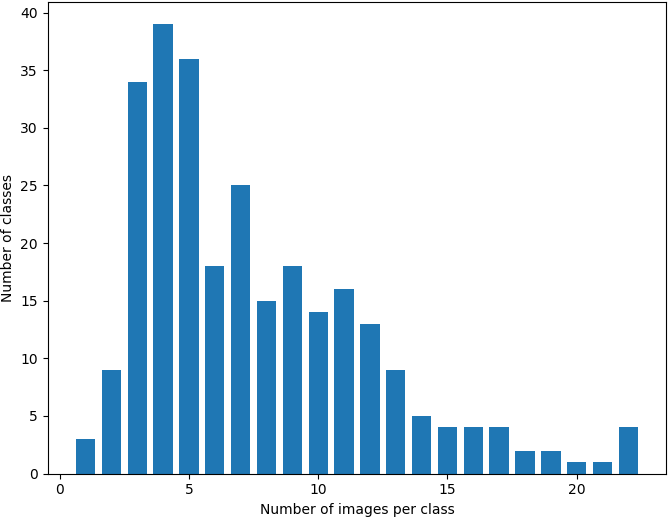}
	\caption{Number of classes with a certain amount of images. The minimum number of images in one class is 1, and the maximum 22.}
	\label{fig:dataset}
\end{figure}

The dataset consists of 2078 images of 276 different elephants. This results in about 8 images per class on average, but as we can see in \autoref{fig:dataset}, the images are not very evenly distributed across the classes. The maximum number of images in a class is 22 and the minimum number is one. We can also see that a big part of the classes only has three to five images, which in turn results in only two to four images for training for these classes.

For our experiments we divided the dataset into a stratified training split of 75\% of the images, and a corresponding test set with 25\%. This results in 1573 images for training and 505 for testing.

\section{Methods}
\label{struc:methods}

\begin{figure}
	\centering
	\includegraphics[width=0.85\linewidth]{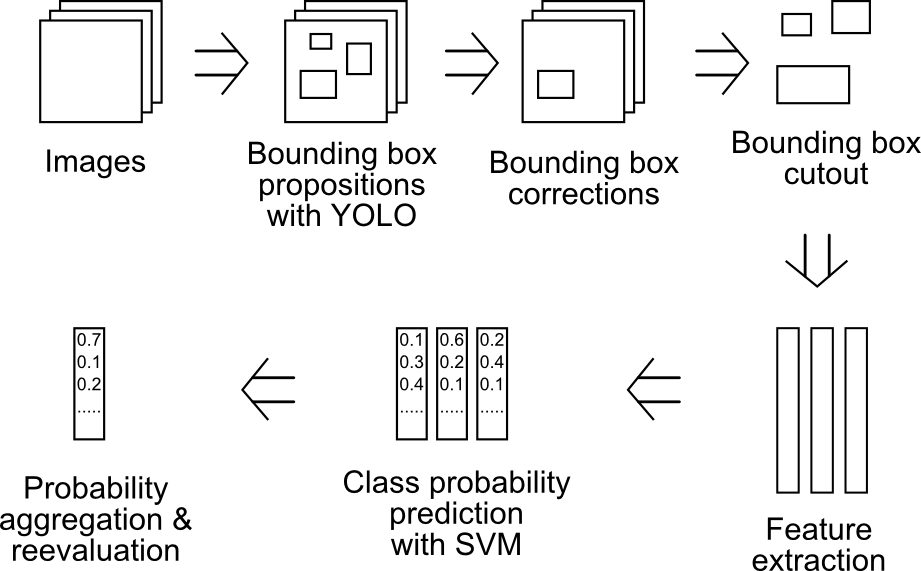}
	\caption{The full pipeline of our system from image input to the classification result.}
	\label{fig:pipeline}
\end{figure}

The processing pipeline of our proposed approach is illustrated in \autoref{fig:pipeline}. Initially, the user inputs one or multiple images. We then automatically locate the elephants' heads in these images using a YOLO network \cite{redmon2016yolo} that has been pre-trained on another dataset built from Flickr images of elephants. This dataset is completely disjoint from the one used for the identification experiments and consists of 1285 training and 227 testing images.

The bounding boxes are drawn around the head instead of the entire body, since preliminary results indicated that the head contains more valuable features for identification.

The predicted bounding boxes are then shown to the user, who then can correct the bounding boxes by drawing a new one, or simply selecting one of multiple proposed ones, as multiple elephants might be contained in the image. 

These selected bounding boxes are then being cut out and fed into a modified ResNet50 network \cite{he2016resnet} for feature extraction. The base network used is the Keras implementation of ResNet50, trained on ImageNet, as the number of images in our dataset is too small to fully train a deep network. It also was modified to extract features not from the last layer before the classification layer, but from earlier activation layers, which are then followed additionally by a new pooling layer to increase translation invariance.

Depending on the layer used for feature extraction and the degree of pooling, the number of features resulting from this may be extremely high, leading to long processing times and possibly memory problems. To deal with this, we apply principal components analysis (PCA \cite{pearson1901pca}) to reduce the number of features to twice the number of training images, \ie, approximately 3,000 for our 75\% training split.

The extracted features are then classified using a support vector machine (SVM \cite{cortes1995svm}) and the classes, \ie, the individual elephants, are sorted in decreasing order by their confidence values obtained from the SVM to create a ranking with the most probable elephant at the top. The ranking is then shown to the user, who can decide, which elephant is the most similar one to the image he provided. A few representative images from the training dataset are shown for each predicted class in the ranking, so that the user can easily filter out false positives.

To leverage the fact that some distinctive features of the elephant head are symmetric, \eg, the tusks or the shape of the ears, we augment the training data for the SVM with features of horizontally flipped versions of all training images.

If multiple images were input, they are used for a joint classification, during which the confidence scores of the SVM for each single image are being aggregated by averaging the class-wise confidence values over all input images. This results in a single confidence score for each individual, so that we can proceed as in the case of a single input image.

\section{Experimental Results}
\label{struc:results}

\begin{table}
    \begin{tabularx}{\linewidth}{|X|c|c|c|c|}\hline
        Top k&1&5&10&20\\\hline
        max\_4 act. 40&0.508&0.706&0.770&0.823\\\hline
        max\_5 act. 40&0.544&\textbf{0.726}&\textbf{0.8}&0.839\\\hline
        max\_6 act. 40&\textbf{0.560}&0.716&0.788&\textbf{0.853}\\\hline
        max\_4 act. 43&0.522&0.716&0.766&0.823\\\hline
        max\_5 act. 43&0.546&0.708&0.770&0.833\\\hline
        max\_6 act. 43&0.524&0.700&0.762&0.821\\\hline
        no pool act. 43&0.518&0.659&0.740&0.805\\\hline
    \end{tabularx}
    \caption{Max pooling with one image using activation\_40 and activation\_43. All pooling trials were done using a network input resolution of $512 \times 512$, the trials without pooling with one of $256 \times 256$. The abbreviations max\_$n$ stand for a max pooling layer with a pooling size of $n \times n$. }
    \label{tbl:pooling_1img}
\end{table}

\begin{table}
    \begin{tabularx}{\linewidth}{|X|c|c|c|c|}\hline
        Top k&1&5&10&20\\\hline
        max\_4 act. 40&0.698&0.818&0.866&0.902\\\hline
        max\_5 act. 40&0.714&0.832&0.876&0.904\\\hline
        max\_6 act. 40&\textbf{0.742}&\textbf{0.852}&\textbf{0.878}&0.906\\\hline
        max\_4 act. 43&0.700&0.830&0.874&\textbf{0.908}\\\hline
        max\_5 act. 43&0.722&0.832&0.876&0.906\\\hline
        max\_6 act. 43&0.708&0.828&0.868&0.904\\\hline
        no pool act. 43&0.686&0.804&0.846&0.886\\\hline
    \end{tabularx}
    \caption{Max pooling with 2 images using the  layers activation\_40 and activation\_43. All pooling trials were done using a network input resolution of $512 \times 512$, the trials without pooling with one of $256 \times 256$. The abbreviations max\_$n$ stand for a max pooling layer with a pooling size of $n \times n$. }
    \label{tbl:pooling_2img}
\end{table}

%Here, we will cover the results regarding object localization and object identification experiments.

\paragraph{Object Localization}

Experiments with the YOLO network suggested that the head of the elephant is easier to detect, and experiments regarding the identification also suggested that head features are preferable for the classification, as they appear to have more features important for identifying the animals. Because of this, we will only focus on the elephant heads and head bounding boxes respectively in the following.

After training the network with 1285 images, its performance was tested on a test set of 227 images, resulting in a precision of 92.73\% and a recall of 92.16\%. The mean average precision achieved was 90.78\%. Thus, we can reliably locate the position of the elephant heads in the images automatically and hence reduce the effort of manual bounding box annotation imposed on the user.

\paragraph{Object Identification}

The identification was done using the combination of a modified ResNet50 as feature extractor and a support vector machine (SVM), which performs the actual classification. The features from the last layer before the classification layer proved to be not the best to extract features from, but the activation layer of the 14$^\mathrm{th}$ residual block in the ResNet50 architecture, here referred to as activation\_43, provided better performance with the final classifier.
This is in line with the findings of \cite{vo2018generalization}, who argue that earlier layers often generalize better to new tasks on datasets different from the training data.

\begin{figure}
	\centering
	\includegraphics[width=.97\linewidth]{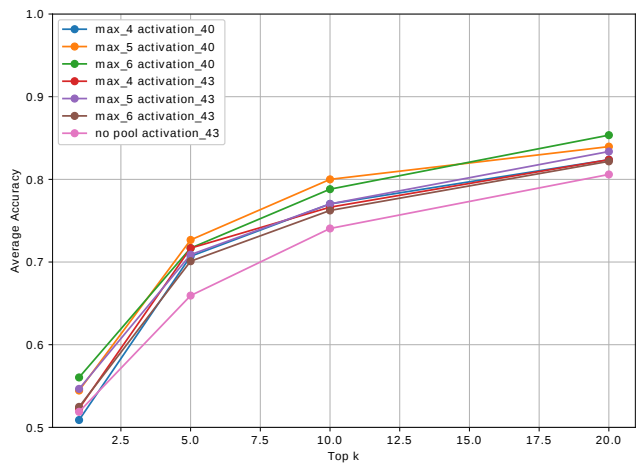}
	\caption{Comparison of different sizes of max pooling layers with single-image classification.}
	\label{fig:pooling_1img}
\end{figure}

If the extracted features are not directly fed into the SVM, but an additional pooling layer is added (cf.\ \autoref{struc:methods}), features extracted from the 13$^\mathrm{th}$ residual block perform even better, as can be seen in \autoref{tbl:pooling_1img} and \autoref{fig:pooling_1img}.
%and perform additional dimensionality reduction as without this, we would have 524288 or even 1048576 features for activation\_43 and activation\_40 respectively. The max pooling with a pooling size of $6 \times 6$ reduces this to 8192 and 25600 respectively. Additionally, we are using a network image input size of $512 \times 512$ and a PCA after feature extraction, which further reduces the number of features to $2 \times sample count$.
%The results using this modified feature extraction and the SVM as classifier can be seen in \autoref{tbl:pooling_1img} and visualized in \autoref{fig:pooling_1img}.
The abbreviations max\_$n$ stand for a max pooling layer with a pooling size of $n \times n$. We can see that the best results using pooling were achieved with the activation\_40 layer, despite the 43rd layer being the best beforehand. The best top-1 accuracy is 56\% with an average per-class accuracy of 49\%. In the top 10 we even achieve up to 80\% and 74\% per-class respectively.

As the dataset is comparatively small compared to most other datasets and most classes have a much smaller number of images than the average of eight, many difficulties can occur. For example, the features needed for correct classification can be missing in the training or testing images, because they are occluded. This can, for example, be caused by mud, or they are simply distorted due to the angle from which the picture was taken. The angle and the movement of the elephant can also have an effect on the features recorded in the picture. Lastly, it could also simply be the case that the elephant to identify was photographed from a view that has never been seen in the dataset, for example an image being input with a left side view of the elephant and the dataset containing only images of the elephant's right side.
In our experiments, we found that a top-1 accuracy of over 70\% can be achieved for elephants with more than 8 training images, whereas the accuracy was below 30\% for individuals with less than 4 images.

\begin{figure}
	\centering
	\includegraphics[width=.97\linewidth]{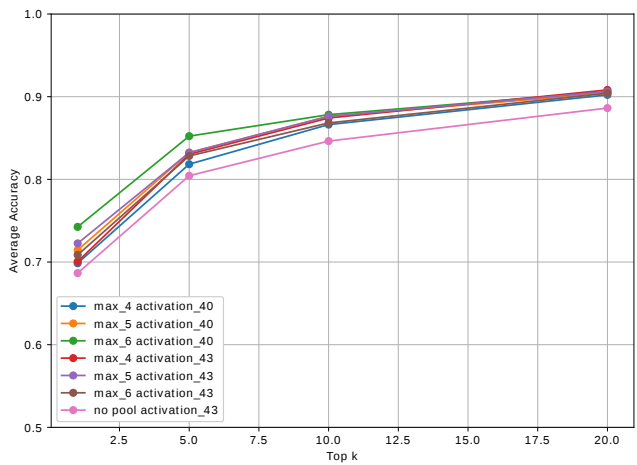}
	\caption{Comparison of different sizes of max pooling layers with two-image classification.}
	\label{fig:pooling_2img}
\end{figure}

All these cases can lead to misclassification. A solution for this problem would be to use multiple images for classification that ideally contain multiple views of the same elephant, and then combine the results of these images as described in \autoref{struc:methods}. The results of this can be seen in \autoref{tbl:pooling_2img} and \autoref{fig:pooling_2img}. We can see that, using two images, an accuracy of 74\% is possible and a per-class average accuracy of 59\%. Among the top 10 results, we even achieve up to 88\% overall and 79\% per-class respectively.

From this we can conclude that it might be a good approach to use multiple images for one classification, as the combined features result in a higher accuracy. This might also be true when using multiple low-quality images, which on their own are not suited for a good classification result, but together it might be possible to still get a successful classification when combining multiple of these.

\section{Problem Analysis}
\label{struc:problems}

As we have seen during the introduction of the dataset used, the images are not distributed very evenly across the classes. This results in many misclassifications, as a lot of classes have only a small number of images to train with. But this is not the only problem we have to deal with in this dataset.

On further analysis of the images we found that some pictures are zoomed in too much, which causes important parts of the elephant's head, like the ears or the tusks, to be outside of the image and thus renders them unusable for classification. In many images the elephants are also very muddy. This often can result in the elephant being mostly monochrome in the image, which is why we cannot see some important contours, \eg, the shape of the ears. Additionally, the images in the dataset have been taken in the years 2000-2016, so a further problem can be that the appearance of the animals and some of their distinctive features changed during that time. This can for example be the case, if the elephant loses a tusk or a hole in the ears becomes a rip.

%Elephants are hard to identify, as they are greatly subject to change due to their long lifetime.

\section{Summary}
\label{struc:summary}

In this project we successfully implemented a system able to assist biologists to identify elephants they encounter in the field. This system consists of a bounding box detector, implemented with YOLO, and a classifier using a combination of a modified ResNet50 as feature extractor, a PCA and an SVM as classifier. These components are connected through a pipeline and can be used via a web interface.

We can achieve about 56\% top-1 and 80\% top-10 accuracy, if we use one image for classification. If we use two images, we can even achieve 74\% and 88\% accuracy respectively. With these results the system will definitely be able to help the biologists in identifying elephants and allow them to focus more on collecting data than on identification.

There are multiple things that still can be done in the project. For example, we noticed that the results are sometimes very dependent on the bounding box drawn. To counter this, an ensemble approach using a multitude of random crops of the original bounding box as input for the classification could be used. Here, a majority vote or also an average of the confidence scores for each crop could be used to receive the actual classification.

%\iffalse
\section*{Acknowledgments}
\label{struc:acknowledgments}

We would like to express our thanks to Andrea Turkalo, who took the images of the elephants used in this work, and the Elephant Listening Project at the Cornell Lab of Ornithology for preparation of the material and metadata.
We would also like to thank Peter Wrege and Daniela Hedwig for their valuable feedback and discussions.
%\fi

\balance
\bibliographystyle{named}
\bibliography{references}

\begin{thebibliography}{}

\bibitem[\protect\citeauthoryear{Brust \bgroup \em et al.\egroup
  }{2017}]{Brust17_TAV}
Clemens-Alexander Brust, Tilo Burghardt, Milou Groenenberg, Christoph Käding,
  Hjalmar Kühl, Marie~L. Manguette, and Joachim Denzler.
\newblock Towards automated visual monitoring of individual gorillas in the
  wild.
\newblock In {\em ICCV Workshop on Visual Wildlife Monitoring (ICCV-WS)}, pages
  2820--2830, 2017.

\bibitem[\protect\citeauthoryear{Cortes and Vapnik}{1995}]{cortes1995svm}
Corinna Cortes and Vladimir Vapnik.
\newblock Support-vector networks.
\newblock {\em Machine learning}, 20(3):273--297, 1995.

\bibitem[\protect\citeauthoryear{He \bgroup \em et al.\egroup
  }{2016}]{he2016resnet}
Kaiming He, Xiangyu Zhang, Shaoqing Ren, and Jian Sun.
\newblock Deep residual learning for image recognition.
\newblock In {\em Proceedings of the IEEE conference on computer vision and
  pattern recognition (CVPR)}, pages 770--778, 2016.

\bibitem[\protect\citeauthoryear{Loos and Ernst}{2013}]{loos2013automated}
Alexander Loos and Andreas Ernst.
\newblock An automated chimpanzee identification system using face detection
  and recognition (cvpr).
\newblock {\em EURASIP Journal on Image and Video Processing}, 2013(1):49,
  2013.

\bibitem[\protect\citeauthoryear{Loos \bgroup \em et al.\egroup
  }{2011}]{loos2011identification}
Alexander Loos, Martin Pfitzer, and Laura Aporius.
\newblock Identification of great apes using face recognition.
\newblock In {\em 19th European Signal Processing Conference}, pages 922--926.
  IEEE, 2011.

\bibitem[\protect\citeauthoryear{Parkhi \bgroup \em et al.\egroup
  }{2015}]{parkhi2015deep}
Omkar~M Parkhi, Andrea Vedaldi, Andrew Zisserman, et~al.
\newblock Deep face recognition.
\newblock In {\em British Machine Vision Conference (BMVC)}, volume~1, page~6,
  2015.

\bibitem[\protect\citeauthoryear{Pearson}{1901}]{pearson1901pca}
Karl Pearson.
\newblock On lines and planes of closest fit to systems of points in space.
\newblock {\em The London, Edinburgh, and Dublin Philosophical Magazine and
  Journal of Science}, 2(11):559--572, 1901.

\bibitem[\protect\citeauthoryear{Redmon \bgroup \em et al.\egroup
  }{2016}]{redmon2016yolo}
Joseph Redmon, Santosh Divvala, Ross Girshick, and Ali Farhadi.
\newblock You only look once: Unified, real-time object detection.
\newblock In {\em Proceedings of the IEEE conference on computer vision and
  pattern recognition (CVPR)}, pages 779--788, 2016.

\bibitem[\protect\citeauthoryear{Schroff \bgroup \em et al.\egroup
  }{2015}]{schroff2015facenet}
Florian Schroff, Dmitry Kalenichenko, and James Philbin.
\newblock Facenet: A unified embedding for face recognition and clustering.
\newblock In {\em Proceedings of the IEEE conference on computer vision and
  pattern recognition (CVPR)}, pages 815--823, 2015.

\bibitem[\protect\citeauthoryear{Taigman \bgroup \em et al.\egroup
  }{2014}]{taigman2014deepface}
Yaniv Taigman, Ming Yang, Marc'Aurelio Ranzato, and Lior Wolf.
\newblock Deepface: Closing the gap to human-level performance in face
  verification.
\newblock In {\em Proceedings of the IEEE conference on computer vision and
  pattern recognition (CVPR)}, pages 1701--1708, 2014.

\bibitem[\protect\citeauthoryear{Turkalo \bgroup \em et al.\egroup
  }{2017}]{turkalo2017slow}
Andrea~K Turkalo, Peter~H Wrege, and George Wittemyer.
\newblock Slow intrinsic growth rate in forest elephants indicates recovery
  from poaching will require decades.
\newblock {\em Journal of Applied Ecology}, 54(1):153--159, 2017.

\bibitem[\protect\citeauthoryear{Vo and Hays}{2018}]{vo2018generalization}
Nam Vo and James Hays.
\newblock Generalization in metric learning: Should the embedding layer be the
  embedding layer?
\newblock {\em arXiv preprint arXiv:1803.03310}, 2018.

\end{thebibliography}

\end{document}